\documentclass[letterpaper, 10 pt, conference]{ieeeconf}  % Comment this line out if you need a4paper
\usepackage{amsmath,amsfonts,amssymb}
\usepackage{algorithmic}
\usepackage{algorithm}
\usepackage{array}
\usepackage[caption=false,font=normalsize,labelfont=sf,textfont=sf]{subfig}
\usepackage{textcomp}
\usepackage{stfloats}
\usepackage{url}
\usepackage{hyperref}
\usepackage{verbatim}
\usepackage{graphicx}
\usepackage{cite}
\usepackage{tabularx}
\usepackage{booktabs}

\IEEEoverridecommandlockouts                              % This command is only needed if 
\title{\LARGE \bf
COP-Q: Safety-First Reinforcement Learning for Robot Control via Cholesky-Ordered Projection
}

\author{Guopeng Li$^{1, 3}$, Moritz A. Zanger$^{2}$, Matthijs T. J. Spaan$^{2}$, Julian F. P. Kooij$^{1}$
\thanks{$^{1}$Department of Cognitive Robotics, Delft University of Technology; $^{2}$Department of Intelligent Systems, Delft University of Technology, the Netherlands; $^{3}$School of Transportation, Southeast University, China}
\thanks{Emails: \texttt{guopengli5@seu.edu.cn}}
\thanks{\texttt{\{m.a.zanger, m.t.j.spaan, j.f.p.kooij\}@tudeflt.nl}}
}

\begin{document}

\maketitle
\thispagestyle{empty}
\pagestyle{empty}

%%%%%%%%%%%%%%%%%%%%%%%%%%%%%%%%%%%%%%%%%%%%%%%%%%%%%%%%%%%%%%%%%%%%%%%%%%%%%%%%
\begin{abstract}

Safe robot control requires maximizing return while satisfying safety constraints. In off-policy safe reinforcement learning, reward and safety Q-values are commonly learned by separate critic ensembles, with uncertainty handled independently for each objective. This objective-wise treatment neglects inter-objective correlation and can lead to overly conservative value estimates, thereby reducing sample efficiency. To address this issue, we propose Cholesky-Ordered Projection Q-learning (COP-Q), a safety-first method that incorporates inter-objective covariance into vector-valued Q-value estimation. COP-Q constructs a generalized confidence bound in the joint Q-value space and uses Cholesky factorization to encode objective priority in a sequential form. This preserves conservatism on safety while adaptively reducing excessive conservatism on the reward objective. The resulting estimate is used in both temporal-difference target computation and actor optimization. COP-Q incurs minimal computational overhead and is readily compatible with most existing deep Q-learning frameworks. Experiments on robot locomotion in Brax and safe navigation in Safety-Gymnasium, covering both hard- and soft-safety settings, demonstrate that COP-Q achieves strong safety performance together with competitive or improved sample efficiency relative to representative baselines.

\end{abstract}

\section{Introduction}

Many robot control tasks require optimizing reward while maintaining safety. For example, legged robot locomotion aims to move faster without falls or hazardous contacts~\cite{yang2022safelegged}, while autonomous vehicles must reach destinations without collisions~\cite{feng2023dense}. Such requirements motivate \textit{safe reinforcement learning} (RL), which aims to maximize cumulative reward while satisfying safety constraints. As RL-based robot control moves closer to deployment, safe RL has received increasing attention~\cite{brunke2022safelearning}.

Deep Q-learning is attractive for safety-critical robot control because its off-policy design and experience replay provide high sample efficiency. However, its effectiveness depends critically on accurate estimation of action-value (Q-value), which is the expectation of discounted cumulative reward or penalty. A central challenge is overestimation bias caused by the maximization step in Bellman updates, which can yield overly optimistic targets and unstable learning~\cite{thrun1993issues}. One common solution is to estimate uncertainty using multiple independent Q-networks and derive conservative value estimates from the ensemble \cite{vanhasselt2016deep}. Such uncertainty-aware methods have been extensively studied in single-objective Q-learning \cite{lockwood2022review}.

In safe RL, however, the agent must estimate both reward and safety Q-values to learn a policy that is both effective and safe. These objectives are typically correlated, so their estimation uncertainties are entangled. Existing methods usually adopt one of two strategies: they either learn reward and safety Q-values independently, as in constrained RL (CRL)~\cite{xu2022constraints}, or handle uncertainty only after scalarizing the objectives, as in multi-objective RL (MORL)~\cite{vanmoffaert2013scalarized}. Although straightforward, both approaches ignore inter-objective correlations and have consequences for safe robot control. 

\begin{figure}[h!]
  \centering
  \vspace{-0.25cm}
\includegraphics[width=0.99\columnwidth]{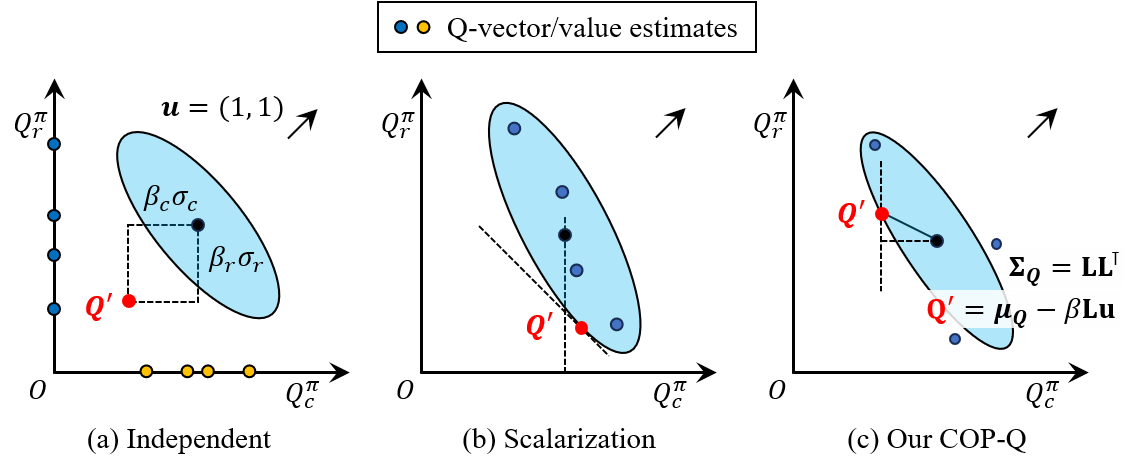}
\vspace{-0.75cm}
  \caption{Obtaining a pair of
  $Q_c^{\pi}$ (safety cost) and $Q_r^{\pi}$ (reward) estimates without overestimation from multiple Q-value predictions by a set of critics. \textit{(a)} Treating two objectives independently may give an over-conservative estimate and reduce sample efficiency, causing a safe policy with a low return; \textit{(b)} Scalarization-based approaches may overestimate safety and lead to an unsafe policy; \textit{(c)} The proposed Cholesky Ordered Projection Q-learning (COP-Q) guides safety-first value-learning by Cholesky factorization, which prioritizes keeping conservatism on safety and uses covariance to adaptively correct over-conservatism of reward.}
  \label{fig: gap}
\end{figure}

Independent modelling combines marginal lower bounds for each objective. When reward and safety Q-values are negatively correlated, this produces an overly conservative estimate and thus causes \textit{low sample efficiency}, as illustrated in Fig. \ref{fig: gap}(a). Scalarization-based uncertainty handling has the opposite weakness: conservatism in the scalarized objective does not guarantee conservatism on safety. Consequently, for negatively correlated objectives, it may still overestimate safety and induce \textit{unsafe behaviours}, as shown in Fig. \ref{fig: gap}(b).

The limitations above highlight a key knowledge gap:

\textit{How can multi-objective uncertainty in vector-valued Q-value estimates be exploited to guide safe yet sample-efficient Q-learning for safety-critical robot control?}

To address this issue, we propose \textit{Cholesky-Ordered Projection Q-learning} (COP-Q). The central idea is to incorporate inter-objective covariance into conservative value estimation while expressing the priority of safety over reward. COP-Q introduces a generalized confidence bound for vector-valued Q-functions and uses Cholesky factorization to adaptively reduce excessive conservatism, as shown in Fig. \ref{fig: gap}(c). This yields a simple and computationally efficient extension of deep Q-learning for safe and sample-efficient robot control.

\section{Related Work}

This section gives a concise overview of the most relevant work to our study, namely safe and constrained RL for robot control, conservative value estimation in deep Q-learning, and vector-valued Q-learning.

Safe robot control is commonly formulated as a constrained Markov decision process (CMDP), where the policy maximizes return subject to safety-cost constraints \cite{wachi2020safe}. Most safe RL methods are on-policy. For example, CPO \cite{achiam2017constrained} provides safety-aware updates with theoretical guarantees, while later methods such as PID-Lagrangian \cite{stooke2020responsive} and CVPO~\cite{liu2022constrained} improve the practicality and stability of constrained policy learning. More recent off-policy safe RL methods further highlight the importance of accurate critic estimation: CAL~\cite{wu2024offpolicy} addresses cost underestimation through conservative cost learning, and ORAC \cite{mccarthy2025optimistic} combines optimistic reward exploration with pessimistic cost estimation. These studies show that safe and sample-efficient control depends critically on reliable reward-cost value estimation. However, they typically learn reward and safety critics separately, or apply uncertainty handling to the cost objective only.

This concern is closely related to the conservative value estimation issue in off-policy deep RL \cite{thrun1993issues}. Double Q-learning~\cite{vanhasselt2010double} and clipped double critics \cite{fujimoto2018addressing} reduce overestimation by constructing more conservative targets, while ensemble methods such as REDQ \cite{chen2021randomized} improve sample efficiency through multiple critics. These methods provide the main algorithmic foundation for conservative TD target construction, but they are developed for single-objective, scalar value functions and therefore do not capture the joint uncertainty structure between reward and safety estimates.

Related issues also arise in multi-objective and vector-valued reinforcement learning \cite{hayes2022practical}. MORL methods such as CAPQL~\cite{lu2023multiobjective} typically learn vector-valued or preference-conditioned Q-functions, together with scalarization mechanisms to handle multiple objectives. While these methods are relevant, their primary goal is preference adaptation or Pareto-oriented learning rather than conservative safety estimation with explicit objective priority. In safety-critical control, however, the safety objective often needs to be monitored explicitly even when a fixed scalarization weight is used, and unsafe events terminate the episode. For example, autonomous driving requires monitoring collision risk \cite{li2026orac} to obtain a safe policy \cite{huang2024safedreamer}. In such settings, mitigating estimation bias in the safety objective is critical.

In summary, existing safe RL and MORL methods either treat reward and safety objectives independently or handle uncertainty only after scalarization. To the best of our knowledge, few deep RL methods explicitly model the joint covariance of reward and safety Q-values. We propose COP-Q to fill this gap.

\section{Problem Formulation}

We consider a bi-objective Markov decision process defined by $(S, A, R, C, p, p_0, \gamma)$. The state space is $S \subseteq \mathbb{R}^m$. For a state $s_t \in S$, a robot controlled by a policy $\pi(\cdot|s)$ takes action $a_t \sim \pi(\cdot|s_t)$ in $A \subseteq \mathbb{R}^d$, the next state follows a distribution $p(s_{t+1} | s_t, a_t)$. The agent receives a reward $r_t$ in $R \subseteq \mathbb{R}$ and a safety signal $c_t$ in $C \subseteq \mathbb{R}$. The initial state distribution is $p_0(s_0)$. The reward and safety action-value functions (Q-functions) are defined as:
\begin{equation}
    \begin{aligned}
    Q_r^{\pi}(s, a) &= \mathbb{E}_{\pi} \left( \sum_{t=0}^{\infty} \gamma^t r_t(s_t, a_t) \mid s_0 = s, a_0 = a \right), \\
    Q_c^{\pi}(s, a) &= \mathbb{E}_{\pi} \left( \sum_{t=0}^{\infty} \gamma^t c_t(s_t, a_t) \mid s_0 = s, a_0 = a \right), 
    \end{aligned}
\end{equation}
where $\gamma \in (0,1]$ is the shared discounting factor. In this study, we consider two types of safe control tasks:

\paragraph{Hard-safety constraint} Unsafe events terminate the episode immediately. The safety signal is a survival reward, and the policy objective is defined as:
\begin{equation}
    Q_{\text{policy}}^{\pi}(s, a) = Q_r^{\pi}(s, a) + Q_c^{\pi}(s, a),
\end{equation}
which gives the aggregate objective. The priority of safety is not imposed by the additive form itself, but is introduced later through the ordered vector-valued Q-function and the safety-first conservative value estimation.

\paragraph{Soft-safety constraint} Unsafe events do not terminate the episode. The agent aims to maximize return under a safety cost constraint, forming a constrained RL (CRL) problem:
\begin{equation}
    \begin{aligned}
    &\max_{\pi} \mathbb{E}_{s\sim\rho_{\pi}, a \sim \pi(\cdot|s)}[ Q_r^\pi (s, a)]\\
    &\text{s.t.} \quad \mathbb{E}_{s\sim\rho_{\pi}, a \sim \pi(\cdot|s)}[ Q_c^\pi (s, a)] \leq d,
    \end{aligned}
\label{eq: cmdp}
\end{equation}
where $\rho_{\pi}$ denotes the state-visitation distribution induced by $\pi$. $d$ is the cost threshold. The primal-dual approach converts the problem into a dual form:
\begin{equation}
    \max \mathbb{E}_{s\sim\rho_{\pi}, a \sim \pi(\cdot|s)}[ Q_r^\pi (s, a) - \lambda(Q_c^\pi (s, a) - d)],
    \label{eq: dual-form}
\end{equation}
\begin{equation}
    \arg\min_{\lambda > 0} \lambda \times (d- \mathbb{E}_{s\sim\rho_{\pi}, a \sim \pi(\cdot|s)}[ Q_c^\pi (s, a)]).
    \label{eq: multiplier}
\end{equation}
The policy $\pi$ and the Lagrangian multiplier $\lambda$ are updated iteratively. $\lambda$ depends on the cumulative safety cost $Q_c^\pi (s, a)$ alone, thus inducing priority between objectives. Note that the safety signal here is a cost penalty, so deep Q-learning needs to mitigate the underestimation bias to ensure conservatism, as explained by Wu et al. \cite{wu2024offpolicy}.

For both tasks, we denote $\mathbf{z}_t = [c_t, r_t]^\intercal$ as the vector of safety and reward signals, and $\mathbf{Q}^{\pi} = [Q^{\pi}_c, Q^{\pi}_r]^\intercal$. The safety objective appears before reward in $\mathbf{Q}^{\pi}$ to indicate its higher priority in value estimation and policy optimization. A corresponding scalarization vector $\mathbf{u} = [u_c, u_r]^\intercal$ can be constant (for hard-safety) or changing (for soft-safety) during training~\cite{wu2024offpolicy}. 

\section{Cholesky ordered projection Q-learning} 
\label{sec: pareto}

We propose the Cholesky ordered projection method, which leverages multi-objective uncertainty estimation to compute a safety-first Q-vector $\mathbf{Q'}$ from an ensemble of critics (Q-networks). To this end, we first extend scalar confidence bounds to vector-valued Q-functions, then propose using Cholesky factorization to impose safety priority without being over-conservative on reward learning. 

% This Q-vector computation is essential for two key steps in Q-learning. One is the next Q-vector $\mathbf{Q'}(s_{t+1}, a')$ in the temporal difference (TD) target $\mathbf{y}_t$ used in critic learning:
% \begin{equation}
%     \mathbf{y}_t = \mathbf{z}_t + \gamma \mathbf{Q'}(s_{t+1}, a'),\ a' = \arg \max_{a} \mathbf{u}^\intercal \mathbf{Q'}(s_{t+1}, a),
%     \label{eq: summation}
% \end{equation}
% where $\mathbf{z}_t = [c_t, r_t]^\intercal$ is the safety-reward signal. The other one is the optimization target for policy optimization:
% \begin{equation}
%     \mathbf{y}_t = \mathbf{z}_t + \gamma \mathbf{Q'}(s_{t+1}, a'),\ a' = \arg \max_{a} \mathbf{u}^\intercal \mathbf{Q'}(s_{t+1}, a),
%     \label{eq: actor update}
% \end{equation}

\subsection{Generalized multi-objective confidence bound}
In single-objective Q-learning, the Q-value estimates given by an ensemble of critics are commonly assumed to be Gaussian distributed \cite{deramo2021gaussian}, denoted as 
$p(Q^{\pi}) = \mathcal{N}(Q^{\pi} | \mu_Q, \sigma^2_Q)$. 
After estimating $\mu_Q$ and $\sigma^2_Q$ from the critics' estimates, a \textit{confidence level} $p \in [0, 1)$ then determines a confidence region on $Q'$ bounded by two endpoints:
\begin{equation}
    \mu_Q \pm \sqrt{\chi^2_1(p)}\sigma_Q,
    \label{eq: single bounds}
\end{equation}
enclosing probability $p$. Here $\chi^2_1(p)$ is the quantile of the Chi-squared distribution. The lower endpoint gives a conservative estimate $Q'=\mu_Q - \sqrt{\chi^2_1(p)}\sigma_Q$ to mitigate overestimation~\cite{vanhasselt2010double}. 

We now generalize this concept to multiple objectives, such as reward and cost, following the same Gaussian assumption. We assume that the Q-vector estimates of critics follow a multivariate Gaussian, denoted as $p(\mathbf{Q}^{\pi}) = \mathcal{N}(\mathbf{Q}^{\pi} | \boldsymbol{\mu_{Q}}, \boldsymbol{\Sigma_{Q}})$. Now a confidence level $p$ determines an \textit{ellipsoid} contour centred at the mean:
\begin{equation}
    (\mathbf{Q}^{\pi} - \boldsymbol{\mu_Q})^\intercal \boldsymbol{\Sigma_Q}^{-1}(\mathbf{Q}^{\pi} - \boldsymbol{\mu_Q}) = \chi^2_N(p),
    \label{eq: ellipsoid}
\end{equation}
where $\chi^2_N(p)$ is the quantile of a multi-dimensional Chi-squared distribution, also known as the squared Mahalanobis distance. To seek a point on the ellipsoid, a possible choice is to apply a linear transformation on the scalarization vector $\mathbf{u}$ with unit norm:
\begin{equation}
    \mathbf{Q'} = \boldsymbol{\mu_Q} + \mathbf{A}\mathbf{u}.
    \label{eq: transform general}
\end{equation}
Inserting $\mathbf{Q'}$ into Eq.~\eqref{eq: ellipsoid} gives the general equation of $\mathbf{A}$, which has infinite solutions. $\boldsymbol{\Sigma_Q}$ admits matrix square-root factorization (we choose positive entries by default):
\begin{equation}
\begin{aligned}
     \boldsymbol{\Sigma_Q} = \mathbf{B} \mathbf{B}^\intercal  \quad &\Rightarrow \quad (\mathbf{A}\mathbf{u})^\intercal (\mathbf{B}^\intercal)^{-1} \mathbf{B}^{-1}(\mathbf{A}\mathbf{u}) = \chi^2_N(p)\\
     &\Rightarrow \quad || \mathbf{B}^{-1}\mathbf{A}\mathbf{u}|| =  \sqrt{\chi^2_N(p)}.
\end{aligned}
\end{equation}
Then the solution has the form:
\begin{equation}
\begin{aligned}
    \mathbf{B}^{-1} \mathbf{A} = \pm \sqrt{\chi_N^2(p)} \mathbf{R}\ \Rightarrow \ &\mathbf{A} = \pm \sqrt{\chi^2_N(p)} \times \mathbf{B} \mathbf{R}\\
    &\text{s.t.}\quad \mathbf{R}^\intercal \mathbf{R} = \mathbf{I},
\end{aligned}
\end{equation}
where $\mathbf{R}$ is an isometry. To preserve the original alignment of the units of objectives, we set $\mathbf{R} = \mathbf{I}$. Then, the conservative projected point on the ellipsoid using the ``-'' sign is:
\begin{equation}
    \mathbf{Q'} = \boldsymbol{\mu_Q} - \sqrt{\chi^2_N(p)} \times \mathbf{B} \mathbf{u}.
    \label{eq: target Q raw}
\end{equation}
This family of solutions is informative because the transformation $\mathbf{B}^{-1}(\mathbf{Q}^{\pi} - \boldsymbol{\mu_Q})$ whitens the Gaussian to a standard form $\mathcal{N}(\boldsymbol{0}, \mathbf{I})$. Conversely, $\mathbf{B}$ re-injects full inter-objective covariances into confidence bounds. Eq.~\eqref{eq: target Q raw} is the \textit{generalized} lower confidence bound for Q-vectors. 

\subsection{Cholesky factorization for safety-first Q-learning}

Note that the factorization of $\boldsymbol{\Sigma_Q}$ is not unique. Every factorization represents a specific way of injecting covariance. As the safety objective has clear priority over reward, we propose using \textit{Cholesky factorization} to reflect this priority. 

The joint distribution $p(\mathbf{Q}^{\pi})$ can be decomposed by the chain rule, and we choose the ordering such that safety appears before reward:
\begin{equation}
    \label{eq: chain-rule}
    p(\mathbf{Q}^{\pi}) = p(\mathbf{Q}_c^\pi) p(\mathbf{Q}_r^{\pi} | \mathbf{Q}_c^{\pi}).
\end{equation}
This motivates an ordered factorization of the covariance structure. Accordingly, we use the Cholesky factorization $\boldsymbol{\Sigma_Q} = \mathbf{L} \mathbf{L}^T$, where $\mathbf{L}$ is the unique lower-triangular factor with positive diagonal entries. The solution becomes:
\begin{equation}
    \mathbf{Q'} = \boldsymbol{\mu_Q} - \sqrt{\chi^2_N(p)} \times\mathbf{L} \mathbf{u}.
    \label{eq: target Q}
\end{equation}
$\mathbf{L}$ induces an ordered correlation structure across objectives. The prioritized safety objective is estimated from its marginal uncertainty only, while the reward objective additionally depends on its covariance with safety. In this way, Eq.~\eqref{eq: target Q} preserves conservatism on safety while using inter-objective correlation to correct excessive conservatism in reward, as illustrated in Fig. \ref{fig: gap}(c). We refer to this method as \textbf{Cholesky-Ordered Projection} (COP).

COP can be used in both critic and actor updates in deep Q-learning methods. This study implements COP into the Soft Actor-Critic framework~\cite{haarnoja2018soft}. When updating critic networks, the mean $\boldsymbol{\hat{\mu}_Q}$ and biased covariance matrix $\boldsymbol{\hat{\Sigma}_Q}$ of the next Q-vector are estimated from the target-critic ensemble. Then we perform Cholesky factorization to get $\mathbf{\hat{L}}(s_{t+1}, a_{t+1})$. If we denote $\beta \equiv \sqrt{\chi^2_N(p)}$, the TD target is:
\begin{equation}
    \begin{aligned}
    \mathbf{y}_t = \mathbf{z}_t + \gamma [&\boldsymbol{\hat{\mu}_Q}(s_{t+1}, a_{t+1}) - \beta \mathbf{\hat{L}}(s_{t+1}, a_{t+1})\mathbf{u}\\
    & - \alpha \mathbf{H}(a_{t+1}|s_{t+1})],\\
    \text{where}\quad &\mathbf{H}(a_{t+1}|s_{t+1}) = [0, \mathcal{H}(a_{t+1}|s_{t+1})]^\intercal.
    \end{aligned}
    \label{eq: td target}
\end{equation}
$\mathcal{H}(a_{t+1}|s_{t+1})$ is the maximum entropy term added to the reward objective. In actor updates, the objective function to maximize follows a similar form:
\begin{equation}
\begin{aligned}
    J^{\pi}_{\text{total}} = \mathbb{E}_{a_t \sim \pi(\cdot|s_t)}[\mathbf{u}^\intercal(&\boldsymbol{\hat{\mu}_Q}(s_t, a_t) - \beta \mathbf{\hat{L}}(s_t, a_t)\mathbf{u})\\
    &- \alpha \mathcal{H}(a_t|s_t)].
\end{aligned}
    \label{eq: actor target}
\end{equation}

The implementation of COP Q-learning (COP-Q) follows a minimal ensemble principle. Estimating the covariance of an $N$-dimensional Q-vector requires at least $N+1$ critics. The biased covariance estimator (dividing by $N+1$) is used due to its smaller mean squared error than the unbiased estimator. This minimal ensemble choice incurs only minimal computational overhead. Also, Cholesky factorization is computationally efficient. The only hyperparameter in COP is the Mahalanobis distance $\beta$, overcoming the difficulty of tuning two hyperparameters for correlated objectives. 

Notably, when $N=1$, COP-Q reduces to the standard scalar conservative target construction, and in the two-critic case, it recovers classical clipped double-Q learning \cite{fujimoto2018addressing}. This reduction shows that COP-Q is a reasonable generalization.

\subsection{Sensitivity to condition numbers}
\label{sec: sensitivity}

COP-Q computes $\mathbf{L}$ from an estimated covariance matrix~$\mathbf{\Sigma}$. So, its numerical stability depends on the conditioning of $\mathbf{\Sigma}$. Following perturbation analysis for Cholesky factorization \cite{chang1996new}, the relative error in $\mathbf{L}$ is bounded by a term proportional to the condition number $\sqrt{\kappa(\mathbf{\Sigma})}$, i.e., the ratio between its largest and smallest eigenvalues:
\begin{equation}
    \dfrac{||\Delta \mathbf{L}||}{||\mathbf{L}||} \leq \sqrt{\kappa(\mathbf{\Sigma})} \dfrac{||\Delta \mathbf{\Sigma}||}{||\mathbf{\Sigma}||}.
\end{equation}
When $\mathbf{\Sigma}$ is ill-conditioned or nearly singular, the Cholesky factor becomes highly sensitive to perturbations in the covariance estimate. This issue is particularly pronounced in two cases: (1) the correlation between objectives is close to $\pm1$, so that the ensemble Q-vectors are nearly collinear in the bi-objective space; (2) the marginal standard deviation ratio $\sigma_c/\sigma_r$ is extreme, e.g., when the uncertainty in safety Q-values is much smaller than that in reward Q-values. In such cases, small estimation errors in $\mathbf{\Sigma}$ can be amplified in $\mathbf{L}$, making the Cholesky-based correction unstable and thereby weakening the practical advantage of COP-Q.

\section{Experiments}

\subsection{Tasks and environments}

\begin{figure}[h!]
  \centering
\vspace{-0.25cm}
\includegraphics[width=\columnwidth]{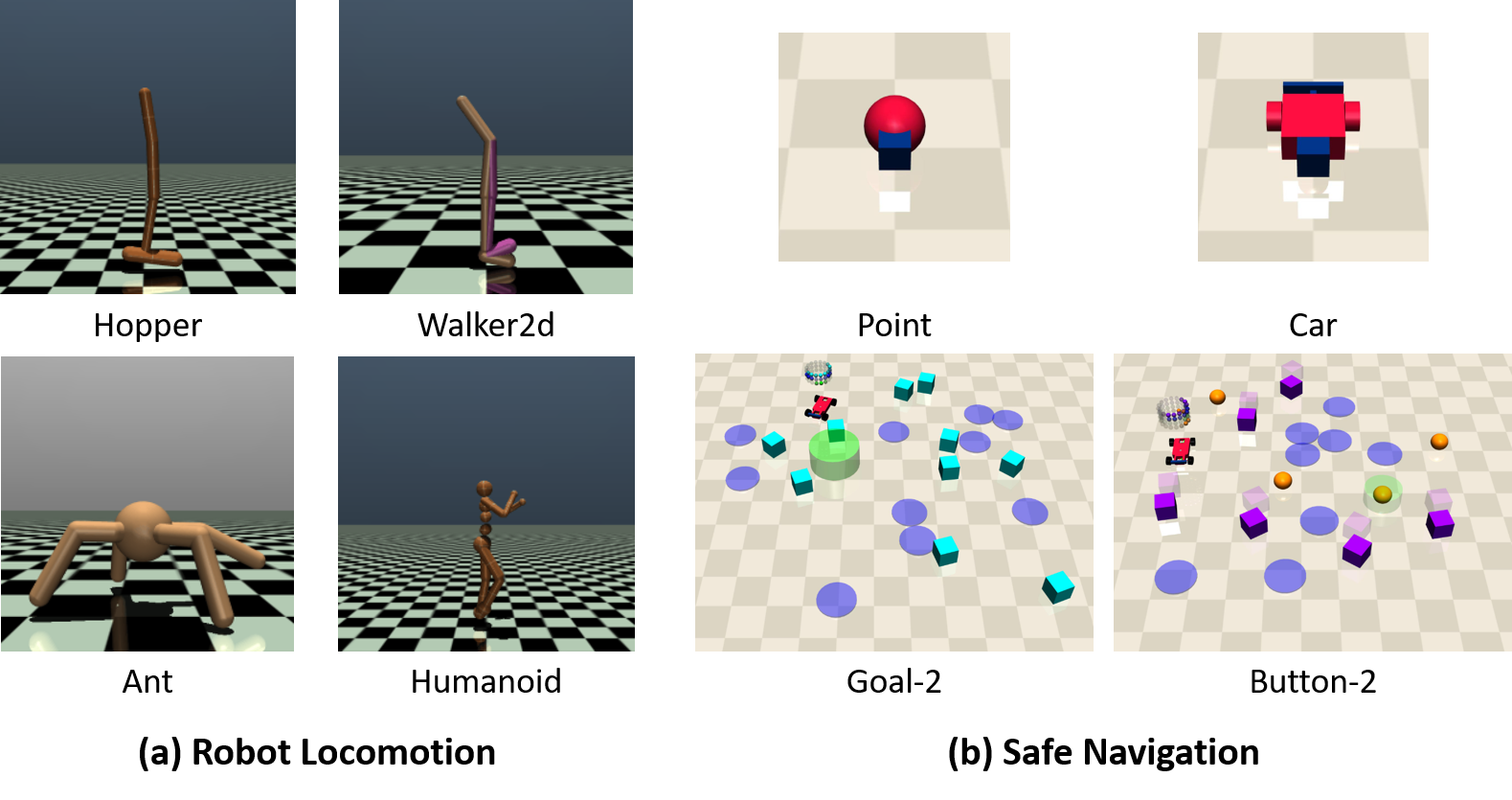}
\vspace{-0.75cm}
  \caption{Illustration of the two experimental environments. (a) Locomotion: 4 selected robot configurations; (b) Safe Navigation: top two panels show the mobile robots used. The bottom two panels show the layout of goals and hazards. For Goal-2, the green object is the only goal position. Blue circles and white cubes are static hazards. For Button-2, there are 3 fake goals and one correct goal position. Besides blue static hazards, the purple cubes are moving gremlins.}
  \label{fig: robot}
\end{figure}

We evaluate COP-Q in two environments: robot locomotion in Brax \cite{freeman2021brax} and Safe Navigation in Safety-Gymnasium \cite{ji2023safetygymnasium}, as shown in Fig. \ref{fig: robot}. The benchmarks include one hard-safety locomotion task (T.1) and two soft-safety constrained RL tasks (T.2 and T.3). They are described as follows.

\paragraph{T.1 Hard-safety robot locomotion}

Four robot configurations in Brax are chosen, as shown in Fig. \ref{fig: robot}(a). Unsafe events terminate the episode immediately. The original robot locomotion task is single-objective. We hereby construct our customized safety and reward signals as follows:
\begin{equation}
c_t = r_{\text{healthy}} + r_{\text{forward}},\quad r_t = - r_{\text{control}}.
    \label{eq: t1_r}
\end{equation}
The healthy reward is +1 if the robot keeps upright, otherwise 0, and terminates the episode. The forward reward is the instantaneous velocity toward a given direction. The control term is the non-negative actuator power, meaning the control effort or energy consumption.

We put the healthy and forward rewards together in the safety objective for two reasons. First, they are strongly aligned. Falling prevents the robot from getting either healthy or forward rewards. Second, the healthy reward alone is binary, and the failure (0) is sparse. This homogeneous signal reduces the dispersion of the estimated safety Q-values, thus causing the stability issue described in Section \ref{sec: sensitivity}.

\paragraph{T.2 Safe Velocity}
Using the same robot configurations, Safe Velocity~\cite{ji2023safetygymnasium} maximizes return while constraining \textit{expected} velocity below a predefined threshold. The signals are defined as:
\begin{equation}
c_t = r_{\text{velocity}},\quad r_t = r_{\text{healthy}} + r_{\text{forward}} - r_{\text{control}}.
    \label{eq: t2_r}
\end{equation}

\paragraph{T.3 Safe Navigation}
We consider Goal2 and Button2 (``2'' indicating the highest level of difficulty) with Point and Car robots on the Safety-Gymnasium \cite{ji2023safetygymnasium}. The robot must reach goals while avoiding static and moving hazards from 2D lidar observations, as illustrated in Fig. \ref{fig: robot}(b). Because Safe Navigation combines sparse safety costs with sparse goal-reaching rewards, it is particularly sensitive to Q-value estimation bias \cite{ji2024omnisafe}. Either overly conservative reward or safety Q-values will significantly suppress the return maximization. 

The maximum episode length is 400 steps, and the cost limit is 10 per episode. COP-Q and baselines are benchmarked on the OmniSafe platform \cite{ji2024omnisafe}. The environment settings are the same as in CVPO \cite{liu2022constrained}.

\subsection{COP-Q implementation and baselines}

In our experiments, each critic network in COP-Q uses a shared trunk with two output heads, corresponding to the reward and safety Q-values. For the hard-safety setting, the raw scalarization vector is fixed as $(1,1)$. In soft-safety CRL, it is updated during training with the Lagrangian multiplier, i.e., $\mathbf{u}=(-\lambda, 1)$. Note that these scalarization vectors must be normalized in COP-Q. To mitigate safety-cost underestimation in CRL, we additionally implement the Augmented Lagrangian Method (ALM) used in CAL \cite{wu2024offpolicy}. 

For the hard-safety locomotion task (T.1), we compare COP-Q with three SAC-based baselines. (1) \textit{Independent} double-Q uses separate critic pairs for the reward and safety objectives. (2) \textit{Scalarization} double-Q uses two critics that jointly predict both Q-values, but selects the critic with the lower scalarized total value for both TD target computation and actor optimization. (3) \textit{Conservative} double-Q uses the same joint critics, but applies the objective-wise lower Q-value in the critic and actor updates.

For the soft-safety CRL tasks (T.2 and T.3), COP-Q is compared against representative off-policy and on-policy baselines. The off-policy baselines are (1) \textit{SACLag} or \textit{SACLag-UCB}~\cite{stooke2020responsive}, which use primal-dual safe RL with single or conservative cost critics, respectively; (2) \textit{CAL}~\cite{wu2024offpolicy}, which combines conservative cost estimation with ALM~\cite{luenberger2016linear}. For the most challenging safe navigation benchmark, we further include (3) \textit{ORAC} \cite{mccarthy2025optimistic}, a recent off-policy baseline that combines quantile-based distributional cost critics with uncertainty-driven exploration. The on-policy baselines are \textit{RCPO} \cite{tessler2019reward}, \textit{CUP} \cite{yang2022constrained}, \textit{PPOSaute} \cite{sootla2022saute}, and \textit{TRPOPID} \cite{stooke2020responsive}.

COP-Q and all off-policy baselines use the same training hyperparameters as in OmniSafe~\cite{ji2024omnisafe}, differing only in the TD target and actor objective computation. We set the update-to-data ratio to 1, without using REDQ \cite{chen2021randomized}. The only COP-Q-specific hyperparameter is Mahalanobis distance $\beta$, which is set to 1 for locomotion (T.1 and T.2) and 0.5 for safe navigation (T.3). Each locomotion experiment is repeated with 10 random seeds. Safe navigation uses 6 seeds due to the longer training time, which is the same as in CAL~\cite{wu2024offpolicy}. 
Additional implementation details are provided in the released code \hyperlink{https://github.com/RomainLITUD/COPQ}{https://github.com/RomainLITUD/COPQ}.

\section{Results and discussion}

\subsection{Hard-safety task (T.1)}

\begin{figure}[h!]
  \centering
  \includegraphics[width=\columnwidth]{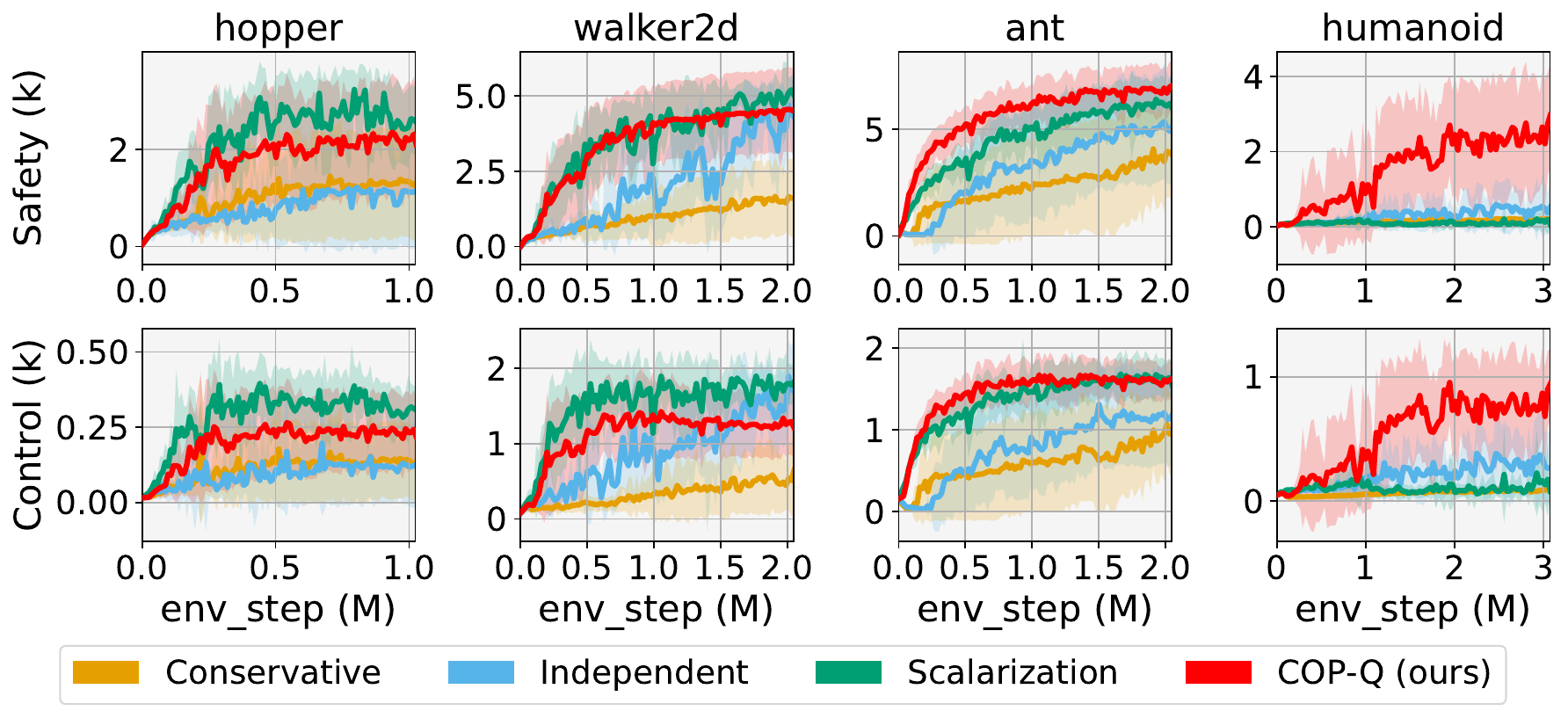}
  \caption{Benchmark on hard-safety robot locomotion tasks. The shaded areas indicate the standard deviation around the mean curve. Same for the following figures.}
  \label{fig: hard-safety}
\end{figure}

The training curves of all methods on T.1 are shown in Fig. \ref{fig: hard-safety}, and the number of falls of the final policy across 1000 test episodes is reported in Table \ref{tab: nb of falls}. COP-Q achieves competitive safety returns and high sample efficiency across the four robot configurations, particularly for the humanoid. In terms of test-time safety, COP-Q yields comparable or fewer falls than the baselines, with its advantage most evident for the humanoid (substantially fewer falls). For the ant, the differences in the number of falls are smaller, which is attributable to the robot's inherently safer dynamics due to its four-legged design.

\vspace{-0.5cm}
\begin{table}[h!]
\caption{The number of falls in 1000 test episodes, Mean (Std.).}
\centering
\begin{tabular}{c|cccc}
\toprule
Method & Hopper & Walker2d & Ant & Humanoid\\
\midrule
Conservative & 19($\pm$14) & 88($\pm$31) & \textbf{21}($\pm$12) & 970($\pm$11)\\
\midrule
Independent & 16($\pm$9) & 19($\pm$10) & 22($\pm$7) & 912($\pm$83)\\
\midrule
Scalarization & 253($\pm$227) & 62($\pm$19) & 31($\pm$16) & 968($\pm$25)\\
\midrule
\textbf{COP-Q (ours)} & \textbf{7}($\pm$6) & \textbf{13}($\pm$6) & 23($\pm$13) & \textbf{31}($\pm$22)\\
\bottomrule
\end{tabular}
\label{tab: nb of falls}
\end{table}

Among the baselines, conservative and independent double-Q learning exhibit low sample efficiency (lower curves on the top row of Fig. \ref{fig: hard-safety}) because their objective-wise lower bound is overly conservative. In contrast, scalarized double-Q shows high sample efficiency for hopper and walker2d, but its resulting policy leads to more falls than COP-Q, especially for the humanoid and hopper. This indicates that conservatism on the scalarized objective alone is insufficient to guarantee conservatism on safety. Taken together, these results support the claim that, in hard-safety tasks, COP-Q preserves safety while improving sample efficiency through adaptive correction of over-conservatism.

\subsection{Soft-safety tasks (T.2 and T.3)}

\begin{figure}[h!]
  \centering
  \includegraphics[width=\columnwidth]{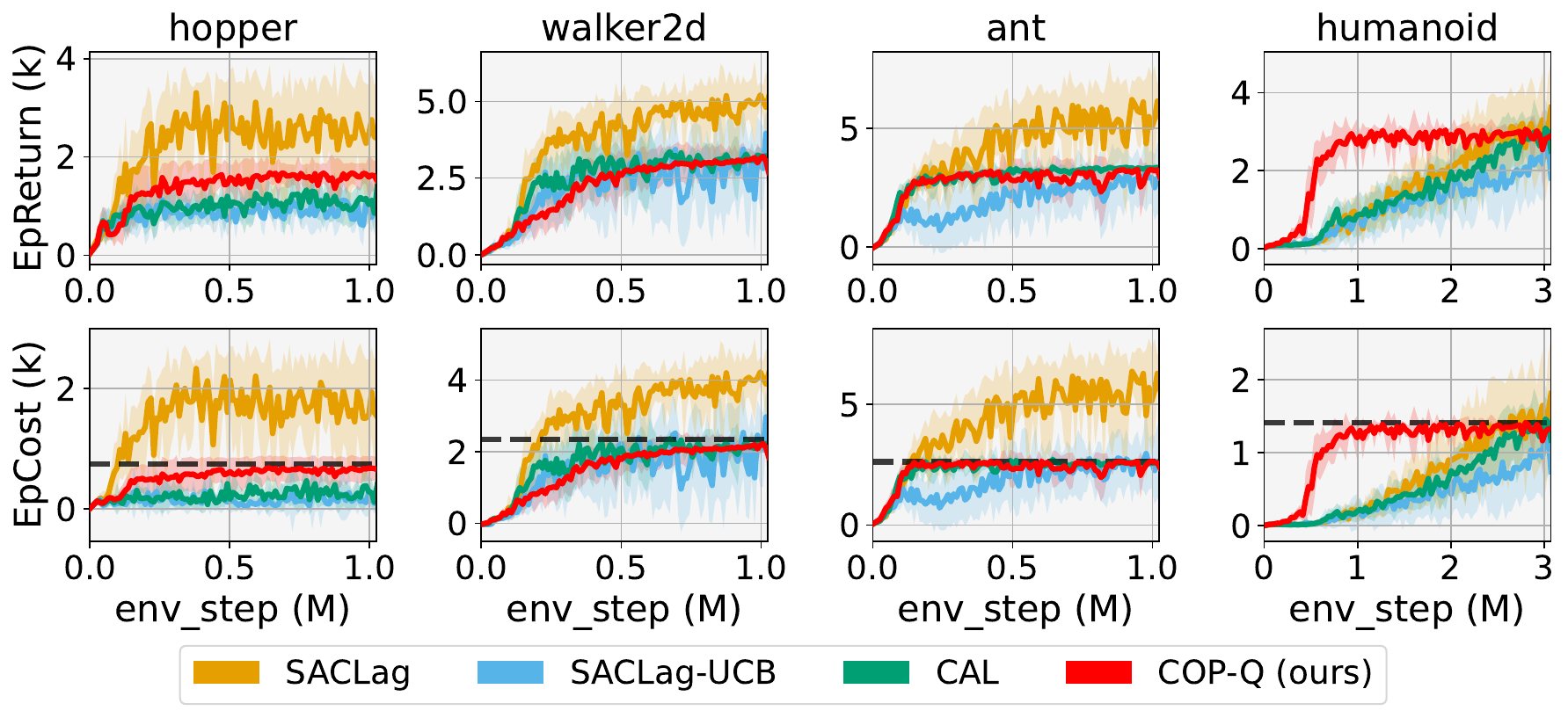}
  \includegraphics[width=\columnwidth]{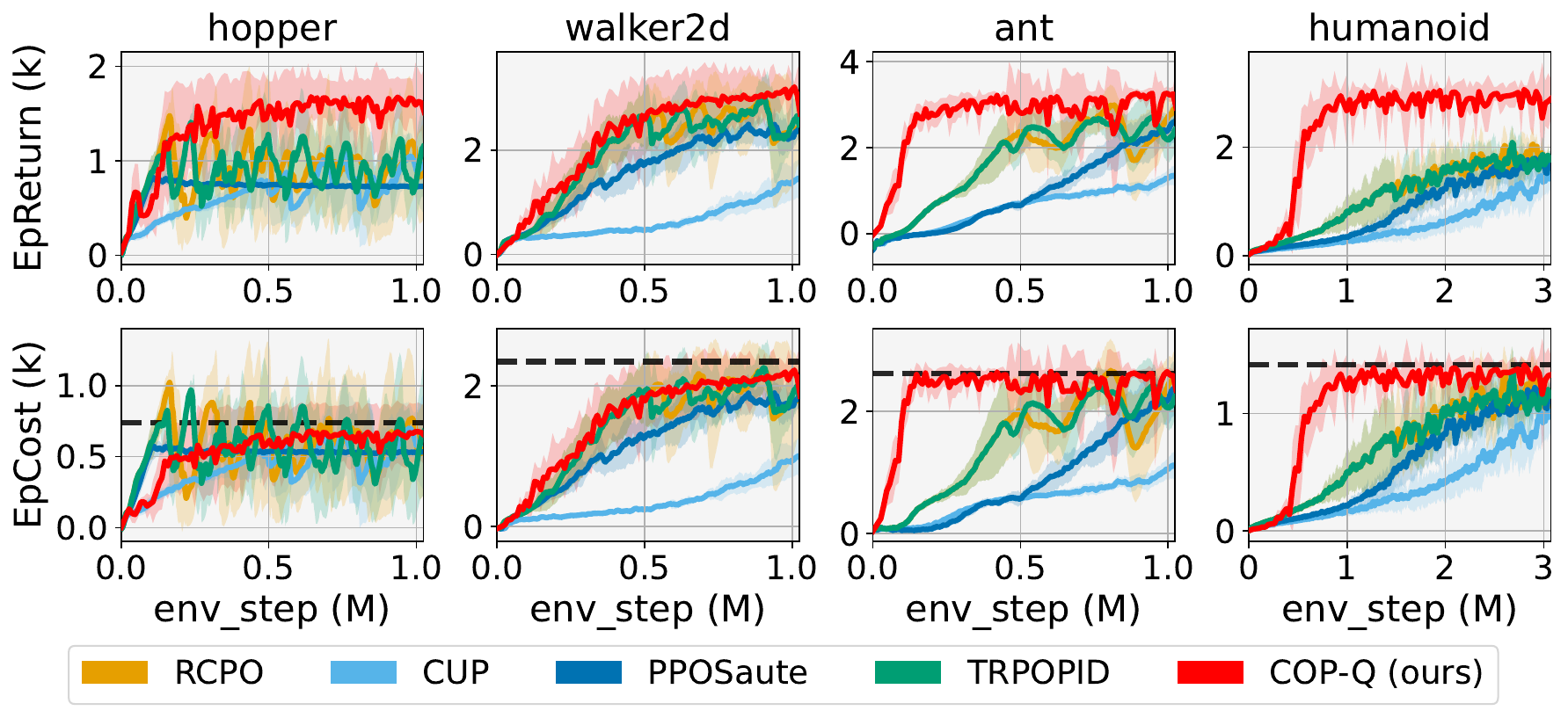}
  \caption{Benchmark on Safe Velocity CRL tasks. COP-Q is compared with off-policy baselines on the top figure and with on-policy baselines on the bottom figure. The black lines mark the upper bound of velocity.}
  \label{fig: safe-vel}
\end{figure}

\begin{figure}[ht!]
  \centering
  \includegraphics[width=\columnwidth]{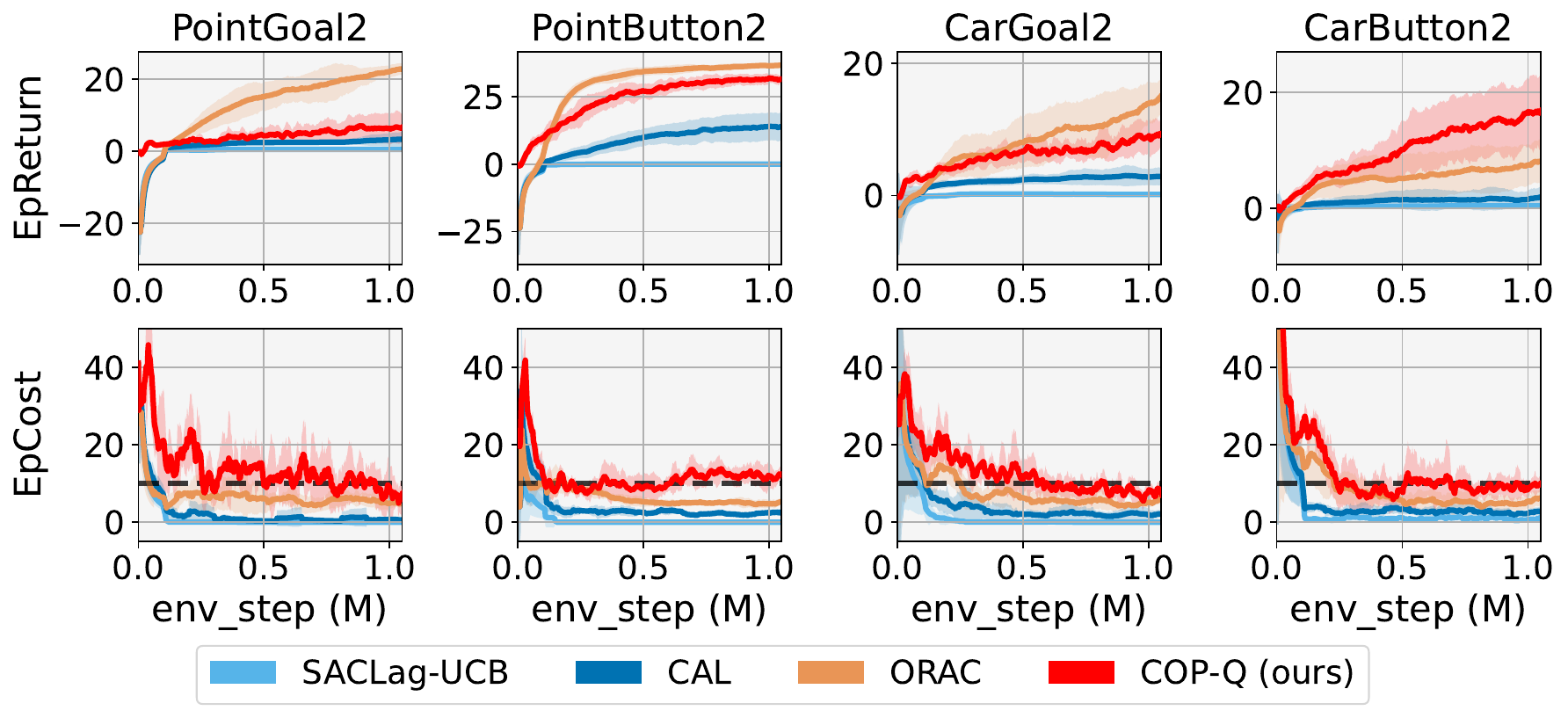}
  \includegraphics[width=\columnwidth]{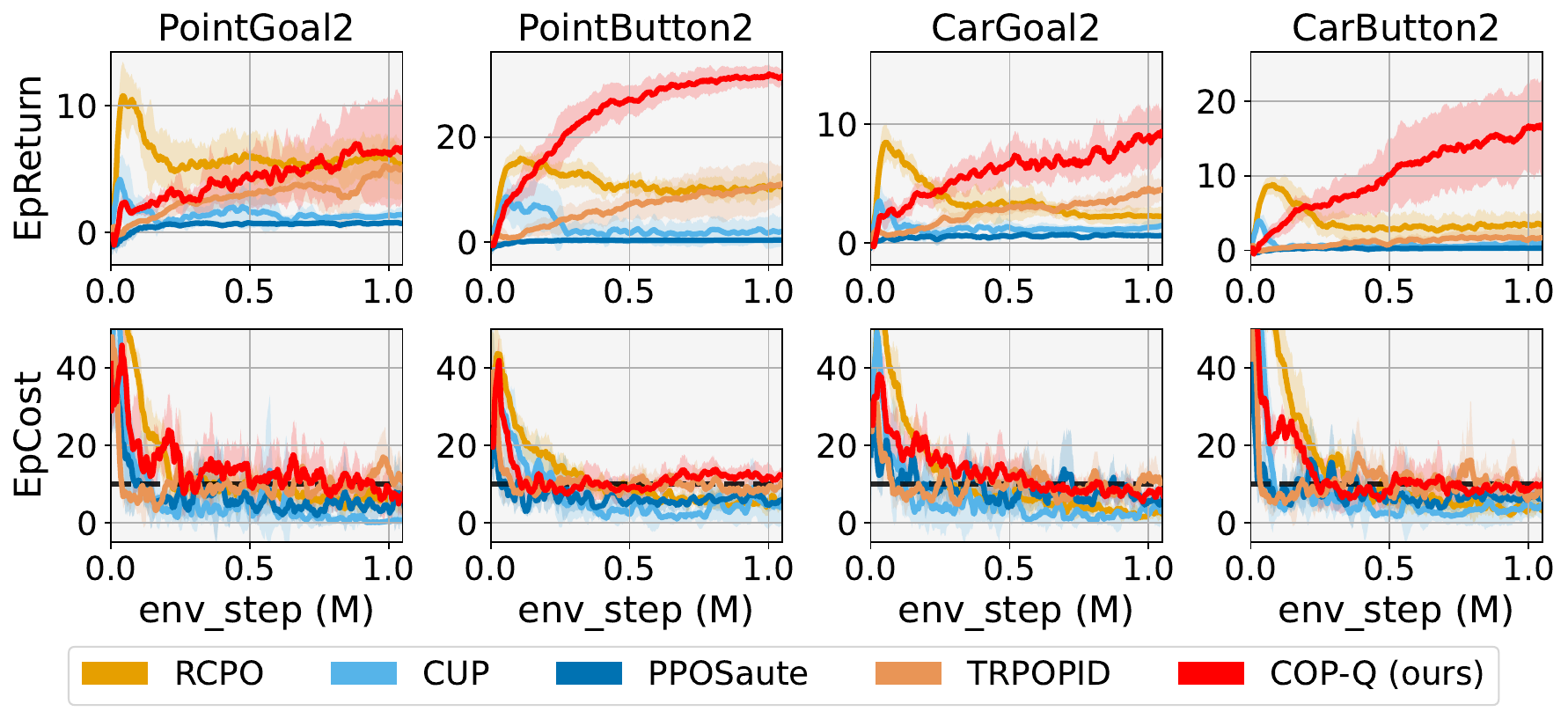}
  \caption{Benchmark on Safe Navigation CRL tasks. COP-Q is compared with off-policy baselines on the top figure and with on-policy baselines on the bottom figure. The black lines mark the upper bound of safety cost.}
  \label{fig: safe-nav}
\end{figure}

The main results on the soft-safety constrained RL tasks are shown in Fig. \ref{fig: safe-vel} (Safe Velocity) and Fig. \ref{fig: safe-nav} (Safe Navigation). Overall, COP-Q maintains the high sample efficiency of off-policy methods while achieving satisfactory safety performance comparable to on-policy methods. In both benchmarks, the cost curve of COP-Q converges below or close to the prescribed thresholds.

\textbf{Safe Velocity:} As shown in Fig. \ref{fig: safe-vel}, SACLag violates the cost constraint (horizontal black dotted lines) on hopper, walker2d, and ant, suggesting that a single cost critic is insufficient to control the underestimation bias of the safety objective. By contrast, SACLag-UCB and CAL produce safer policies, but at the cost of lower sample efficiency (see the lower EpReturn curves), especially on hopper and humanoid. COP-Q alleviates this trade-off by incorporating inter-objective covariance into safety-first value estimation, thereby improving return while avoiding any sacrifice in safety performance.

\textbf{Safe Navigation:} The same tendency becomes even more pronounced in Fig. \ref{fig: safe-nav}, as this benchmark is more sensitive to Q-value estimation bias. SACLag-UCB and CAL converge to overly conservative policies with nearly zero cost but substantially lower return than COP-Q. This indicates that independent conservative estimation can suppress efficient reward learning in challenging safe RL tasks. By contrast, COP-Q preserves conservatism on the safety objective while adaptively reducing excessive conservatism on the reward objective, leading to safe policies (the EpCost curves constrained below the threshold in Fig. \ref{fig: safe-nav}) that reach goals more effectively than the baselines (the higher EpReturn in Fig. \ref{fig: safe-nav}). Moreover, the return of COP-Q is comparable to that of the state-of-the-art method ORAC in PointButton2, CarGoal2, and CarButton2, even though ORAC relies on additional advanced techniques such as optimistic exploration and quantile-based distributional cost estimation.

In summary, the results on Safe Velocity and Safe Navigation highlight the central advantage of COP-Q for soft-safety constrained RL: by modeling inter-objective covariance, COP-Q improves sample-efficient policy learning without sacrificing safety. This benefit is especially pronounced in tasks where independent conservative estimation—used by methods like SACLag-UCB and CAL—becomes overly pessimistic, suppressing reward-efficient exploration. Thus, COP-Q offers a principled balance between safety and reward optimization where prior approaches either violate constraints or converge to overly conservative solutions.

\subsection{Analysis on correlation between objectives}

The benefit of COP-Q is expected to be greater when the reward and safety objectives are more negatively correlated, since independent conservative estimation becomes more likely to produce overly pessimistic targets in this regime. To examine this effect, we compute the average Pearson correlation coefficient $\rho \in [-1, 1]$ during training before COP-Q reaches 80\% of its highest episode return. The results for the three tasks are summarized in Table \ref{tab: correlation}.

For the locomotion tasks, humanoid has the most negative correlation among the four robots in both T.1 and T.2, while walker2d and ant are close to zero. This is consistent with the empirical results in Fig. \ref{fig: hard-safety} and \ref{fig: safe-vel}, where the advantage of COP-Q is more significant for humanoid and relatively marginal for walker2d and ant. A similar pattern is observed in Safe Navigation T.3, where the correlation coefficients are mostly near zero or negative. This is again consistent with the gains of COP-Q over independently conservative baselines such as SACLag-UCB and CAL. 

\begin{table}[!ht]
\caption{Average Pearson correlation coefficient $\rho$, P- and C- are abbreviations for Point and Car.}
\centering
\begin{tabular}{c|cccc}
\toprule
Hard-Safety T.1 & Hopper & Walker2d & Ant & Humanoid\\
\midrule
$\rho$ & -0.13 & -0.07 & -0.05 & -0.41\\
\midrule
Safe Vel. T.2 & Hopper & Walker2d & Ant & Humanoid\\
\midrule
$\rho$ & -0.27 & 0.11 & 0.04 & -0.39\\
\midrule
Safe Nav. T.3 & P-Goal2 & P-Button2 & C-Goal2 & C-Button2\\
\midrule
$\rho$ & -0.09 & -0.15 & -0.17 & -0.25\\
\bottomrule
\end{tabular}
\label{tab: correlation}
\end{table}

The correlation analysis supports the central intuition of COP-Q: incorporating inter-objective covariance is particularly useful when the reward and safety objectives are negatively correlated, because it can reduce unnecessary conservatism while preserving safety.

\subsection{Ablations on binary sparse signals for T.1}

To examine a challenging failure mode of COP-Q, we conduct an ablation experiment for the hard-safety task T.1 by defining a binary sparse safety signal with continuous dense reward:
\begin{equation}
c_t = r_{\text{healthy}},\quad r_t = r_{\text{forward}} - r_{\text{control}}.
    \label{eq: t1_re}
\end{equation}
This definition significantly reduces the dispersion of the estimated safety Q-values across critics. As a result, the covariance matrix of the vector-valued Q-estimates is more likely to be ill-conditioned, which is the challenging case discussed in Section \ref{sec: sensitivity}. 

\begin{figure}[ht!]
  \centering
  \includegraphics[width=\columnwidth]{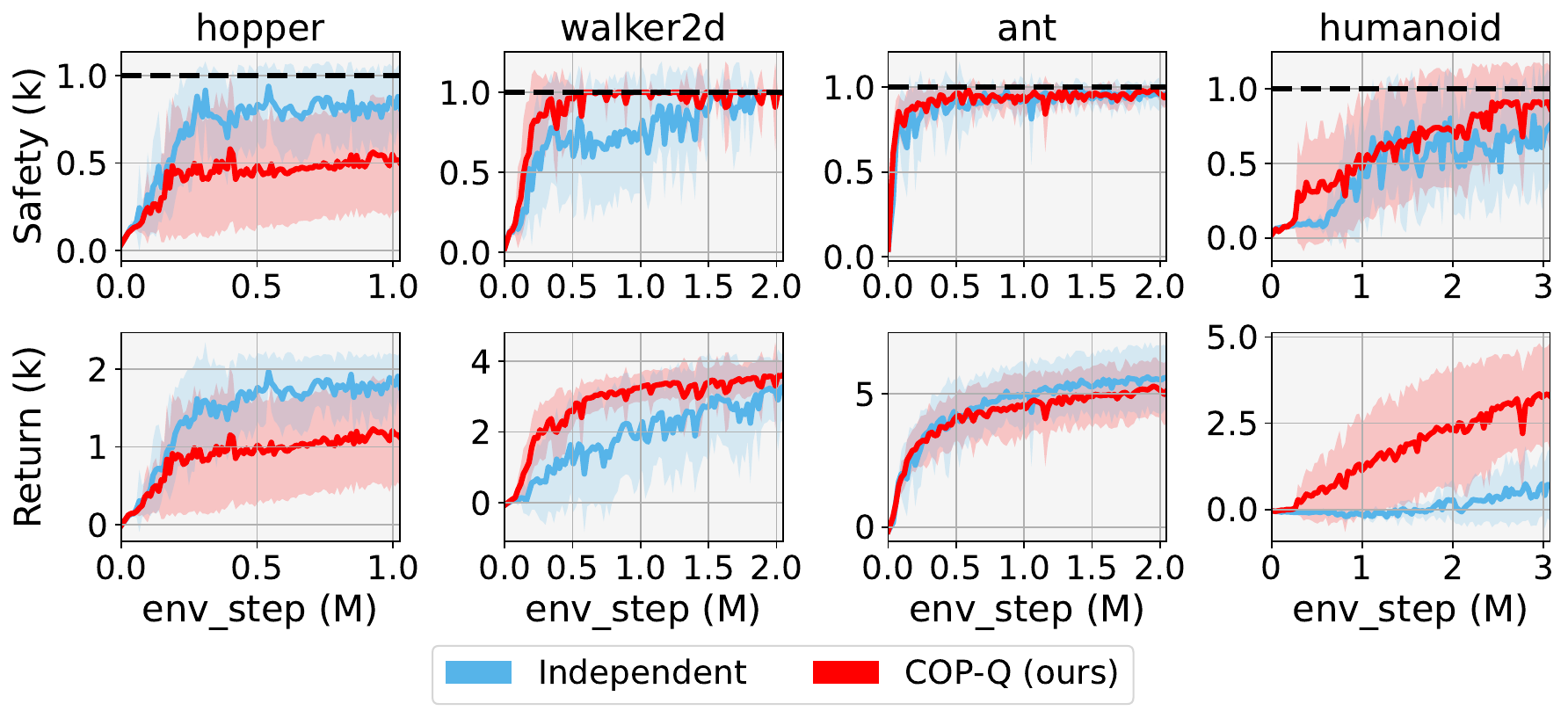}
  \caption{Comparison between the independent double-Q baseline and COP-Q for binary sparse survival rewards. The black lines mark the maximum possible episode safety return, corresponding to 1000 steps without falls.}
  \label{fig: ablation}
\end{figure}

The experimental results are presented in Fig. \ref{fig: ablation}. Under this setting, COP-Q does not demonstrate an advantage over the independent double-Q baseline with respect to safety performance. The training curves of COP-Q are particularly unstable for hopper and humanoid, as indicated by the wide shaded area representing standard deviation. This ablation highlights an important limitation of COP-Q: when the safety signal is binary, sparse, and nearly homogeneous, the covariance matrix can be ill-conditioned, and thus degrade learning stability.

\section{Conclusions}

This paper presented Cholesky-Ordered Projection Q-learning (COP-Q), a safety-first RL method for conservative value estimation in safe robot control. COP-Q incorporates inter-objective covariance into vector-valued Q-learning and uses Cholesky factorization to encode the priority of safety over reward. This yields a simple extension of deep Q-learning that preserves conservatism on the safety objective while adaptively reducing excessive conservatism on the reward objective.
Experiments on robot locomotion and safe navigation benchmarks showed that COP-Q achieves a favourable balance between safety and sample efficiency across both hard- and soft-safety settings.

\textbf{Limitations:} COP-Q relies on stable covariance estimation and can become unreliable when the covariance matrix is ill-conditioned, especially under binary and sparse safety signals. In addition, the current formulation captures total estimation uncertainty without distinguishing epistemic and aleatoric components, and its applicability to broader multi-objective settings remains to be studied. These limitations motivate future work on more robust covariance estimation, richer uncertainty modelling, and extensions to more general multi-objective reinforcement learning problems.

%%%%%%%%%%%%%%%%%%%%%%%%%%%%%%%%%%%%%%%%%%%%%%%%%%%%%%%%%%%%%%%%%%%%%%%%%%%%%%%%



% \section*{ACKNOWLEDGMENT}

%%%%%%%%%%%%%%%%%%%%%%%%%%%%%%%%%%%%%%%%%%%%%%%%%%%%%%%%%%%%%%%%%%%%%%%%%%%%%%%%

\bibliographystyle{IEEEtran}
\bibliography{ref}

\end{document}